\pdfoutput=1

\documentclass[11pt]{article}
\usepackage{tikz}
\usetikzlibrary{positioning, shapes, arrows.meta, calc}
\usepackage{caption}
\usepackage[final]{mtsummit25}
\usepackage{hyperref}
\usepackage{amsmath}
\usepackage{times}
\usepackage{latexsym}
\usepackage{graphicx}
\usepackage{booktabs}
\usepackage{pgfplots}
\usepackage{pgfplotstable}
\usepackage[T1]{fontenc}
\usepackage{multirow}
\usepackage{xcolor}
\usepackage{adjustbox}

\usepackage[utf8]{inputenc}

\usepackage{microtype}

\usepackage{inconsolata}

\usepackage{graphicx}

\title{Extending CREAMT: Leveraging Large Language Models\\ for Literary Translation Post-Editing}

\author{Antonio Castaldo \\
  University of Naples ``L'Orientale'' \\
  University of Pisa \\
  \texttt{antonio.castaldo@phd.unipi.it} \\\And
  Sheila Castilho \\
  Dublin City University \\
  \texttt{sheila.castilho@adaptcentre.ie} \\\AND
  Joss Moorkens\\
  Dublin City University\\ 
  \texttt{joss.moorkens@dcu.ie} \\\And
  Johanna Monti\\
  University of Naples ``L'Orientale''\\
  \texttt{jmonti@unior.it} \\
  }

\begin{document}
\maketitle

\begin{abstract}

Post-editing machine translation (MT) for creative texts, such as literature, requires balancing efficiency with the preservation of creativity and style. While neural MT systems struggle with these challenges, large language models (LLMs) offer improved capabilities for context-aware and creative translation. This study evaluates the feasibility of post-editing literary translations generated by LLMs. Using a custom research tool, we collaborated with professional literary translators to analyze editing time, quality, and creativity. Our results indicate that post-editing LLM-generated translations significantly reduces editing time compared to human translation while maintaining a similar level of creativity. The minimal difference in creativity between PE and MT, combined with substantial productivity gains, suggests that LLMs may effectively support literary translators working with high-resource languages.

\end{abstract}

\section{Introduction}

Post-editing of MT has become an increasingly common service, given the cost-efficiency and good quality compromise that this practice offers. However, while several studies have confirmed that post-editing MT boosts productivity in terms of translation speed \cite{terribile_is_2023}, the benefits diminish significantly when dealing with poor-quality MT outputs \cite{guerberof_arenas_correlations_2014,sanchez-torron_machine_2016}. This challenge is particularly pronounced for literary texts, where the final quality often suffers not only in terms of translation accuracy but also in the preservation of creativity, as discussed by \citet{guerberof-arenas_impact_2020}.

Recent LLM advancements have demonstrated significant improvements in handling context issues and figurative language to generate highly accurate and fluent translations. Unlike NMT systems that often tend towards generating translations that are either too literal or inaccurate, LLMs leverage large training data to generate context-aware translations less literally. Nevertheless, the extent to which they may support literary translators, without sacrificing creativity, remains underexplored.

In this study, we collaborated with four professional translators to evaluate the feasibility of post-editing literary translations generated by LLMs, focusing on three key aspects: editing time, translation quality, and creativity. We compare the performance of GPT-4, GPT-3.5, and a literary-adapted Mistral-7B model. We also developed a custom research tool called UniOr-PET \citep{castaldo_2025_UniOrPET} to collect detailed statistics on the editing process of a literary sci-fi novel.

Our findings reveal that post-editing LLM-generated translations between well-supported languages significantly reduces editing time compared to human translation while maintaining a similar level of creativity. As the difference in creativity scores between human and post-edited LLM translations appears to be minimal, our findings suggest that LLMs can serve as valuable tools for literary translators.

\section{Related Work}

Research on post-editing has traditionally centered on technical and commercial texts, where terminological consistency and turnaround time are often prioritized \citep{moorkens_translators_2018}. However, translating creative works such as literature introduces unique challenges. NMT models have been shown to struggle with creative phraseological challenges, such as translating idiomatic expressions, where they often produce overly literal outputs.

\citet{corpas_pastor_human_2024} highlighted these limitations, particularly in the context of literary texts. In contrast, \citet{raunak_gpts_2023} demonstrated that LLMs are capable of generating less literal and more contextually appropriate translations, especially when translating idiomatic expressions that tend to be generated with a higher level of abstraction, defined by the authors as ``figurative compositionality''. Further studies on idiomatic expression translation, particularly for the English-Italian language pair, have confirmed the high-quality results achieved by general-purpose LLMs \citep{castaldo_prompting_2024}. Their findings suggest that these models could address some of the shortcomings observed in NMT systems when translating literature, making them a promising tool for literary translation.

A study conducted by \citet{guerberof-arenas_creativity_2022} concluded that NMT was unable to handle the complex demands of translating literature or supporting literary translators effectively, resulting in low-quality outputs and diminished creativity. Their findings revealed the limitations of such models in preserving creativity during translation, becoming a constraint for the translator's creativity when used. Human translation (HT) consistently outperformed MT and PE in creativity, as evidenced by the annotation of units of creative potential. These findings align with the study by \citet{castilho_post-editese_2022}, that showed how the features found in post-edited translations align more closely with the ones found in the MT output than in the HT. However, more recent advances in LLMs may shift this paradigm.

As demonstrated by \citet{karpinska_large_2023} and \citet{castilho_online_2023}, LLMs excel at leveraging training data to deal with context-related issues, which is critical for translating creative works that require discourse-level coherence and contextual understanding. Techniques such as in-context learning \citep{brown_language_2020} and prompt engineering allow LLMs to maintain higher degrees of fluency, consistency, and stylistic fidelity compared to NMT systems. Finally, their ability to adapt to specific linguistic patterns and translation memories in real time, as shown by \citet{moslem_adaptive_2023}, further enhances their applicability in the creative translation domain, suggesting that LLMs could potentially overcome the creativity gap identified in NMT outputs, supporting professional translators in producing high-quality creative translations with context-aware terminology and accurate lexicon.

Drawing on \citet{guerberof-arenas_creativity_2022}, in this study we consider creativity as a process that requires both originality and effectiveness \citep{runco_standard_2012}. This implies that in order for a product to be creative, it needs not only to be novel but also of value, and therefore acceptable, for the context in which it is created. In Section 5, we will use the annotations of units of creative potential to reflect the original units introduced by the translators  (novelty), and translation quality metrics as a proxy for the translation acceptability.

\section{Methodology}

We collaborated with four professional translators who specialize in literary and editorial translations to translate and post-edit excerpts from the novel ``Oryx and Crake'' by Canadian author Margaret Atwood \citep{atwood_oryx_2004} from English into Italian. The novel was selected for its extensive use of playful and thought-provoking neologisms, vivid imagery, and richly detailed language, which present significant challenges in the translation process \citep{miller_punsters_2019,gurov_literary_2022,noriega-santianez_machine_2023}

\subsection{Participants}

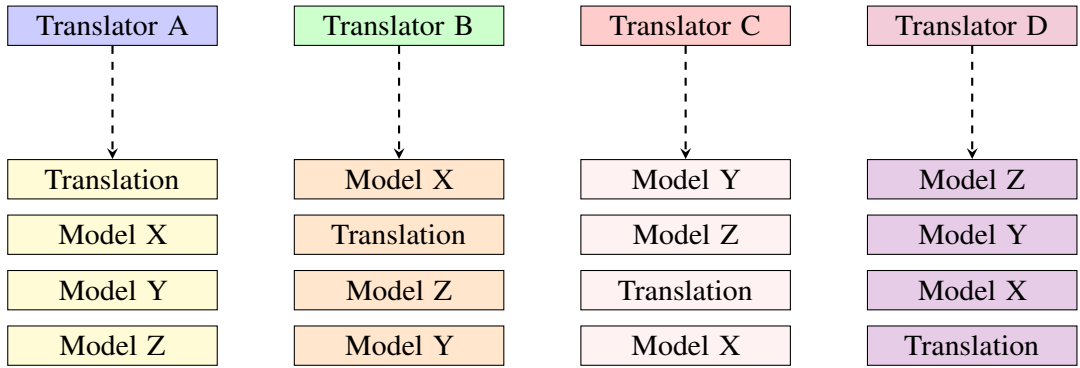
\begin{figure*}[!h]
    \centering
    \begin{tikzpicture}[>=stealth, node distance=1.2cm]

\node[draw, rectangle, fill=blue!20, text width=2.5cm, align=center] (A) {Translator A};
\node[draw, rectangle, fill=green!20, text width=2.5cm, align=center, right=1cm of A] (B) {Translator B};
\node[draw, rectangle, fill=red!20, text width=2.5cm, align=center, right=1cm of B] (C) {Translator C};
\node[draw, rectangle, fill=purple!20, text width=2.5cm, align=center, right=1cm of C] (D) {Translator D};

\path (B) -- (C) coordinate[midway] (BC);

\node[draw, rectangle, fill=yellow!20, text width=2.5cm, align=center, below=1.5cm of A] (X1) {Translation};
\node[draw, rectangle, fill=yellow!20, text width=2.5cm, align=center, below=0.2cm of X1] (X2) {Model X};
\node[draw, rectangle, fill=yellow!20, text width=2.5cm, align=center, below=0.2cm of X2] (X3) {Model Y};
\node[draw, rectangle, fill=yellow!20, text width=2.5cm, align=center, below=0.2cm of X3] (X4) {Model Z};

\node[draw, rectangle, fill=orange!20, text width=2.5cm, align=center, below=1.5cm of B] (Y1) {Model X};
\node[draw, rectangle, fill=orange!20, text width=2.5cm, align=center, below=0.2cm of Y1] (Y2) {Translation};
\node[draw, rectangle, fill=orange!20, text width=2.5cm, align=center, below=0.2cm of Y2] (Y3) {Model Z};
\node[draw, rectangle, fill=orange!20, text width=2.5cm, align=center, below=0.2cm of Y3] (Y4) {Model Y};

\node[draw, rectangle, fill=pink!20, text width=2.5cm, align=center, below=1.5cm of C] (Z1) {Model Y};
\node[draw, rectangle, fill=pink!20, text width=2.5cm, align=center, below=0.2cm of Z1] (Z2) {Model Z};
\node[draw, rectangle, fill=pink!20, text width=2.5cm, align=center, below=0.2cm of Z2] (Z3) {Translation};
\node[draw, rectangle, fill=pink!20, text width=2.5cm, align=center, below=0.2cm of Z3] (Z4) {Model X};

\node[draw, rectangle, fill=violet!20, text width=2.5cm, align=center, below=1.5cm of D] (W1) {Model Z};
\node[draw, rectangle, fill=violet!20, text width=2.5cm, align=center, below=0.2cm of W1] (W2) {Model Y};
\node[draw, rectangle, fill=violet!20, text width=2.5cm, align=center, below=0.2cm of W2] (W3) {Model X};
\node[draw, rectangle, fill=violet!20, text width=2.5cm, align=center, below=0.2cm of W3] (W4) {Translation};

\draw[->, thick, dashed] (A) -- (X1);
\draw[->, thick, dashed] (B) -- (Y1);
\draw[->, thick, dashed] (C) -- (Z1);
\draw[->, thick, dashed] (D) -- (W1);

\end{tikzpicture}
    \caption{Each translator translates from scratch one chunk of original text (Translation) and post-edits a different chunk of each model's output (Model X, Y, Z), minimizing the translator's effect.}
    \label{fig:exp-setup}
\end{figure*}

Each translator post-edited outputs of comparable length (roughly 2200 words), generated by three LLMs (see~§\ref{subsec:models-and-training}). We designed our study so that each translator contributes equally to the evaluation of the four models, rotating the chunks so that each translator works on three unique chunks, each generated by a different model. In this way, we minimize biases introduced by translator-specific behavior. We demonstrate our approach in Figure~\ref{fig:exp-setup}.

In addition, each translator produced a segment of the same excerpt translated from scratch. This experimental setup enabled us to collect fully post-edited translations for each model and a complete HT of the text for comparative analysis.

\subsection{Models and Training}
\label{subsec:models-and-training}

We employed three LLMs for generating the initial translations: \texttt{GPT-4}, \texttt{GPT-3.5}, and a literary-adapted \texttt{Mistral-7B} model, ordered by parameter size. Access to the GPT models \citep{openai_gpt-4_2024} was obtained through the OpenAI API,\footnote{\url{https://openai.com}} as they both operate under closed-source licenses. In contrast, Mistral-7B \citep{jiang_mistral_2023} was obtained as an open-source checkpoint, allowing us to fine-tune it locally for literary translation. Mistral-7B was fine-tuned on a curated corpus of modern literary works obtained from Opus Corpus \citep{tiedemann_opus-mt_2020}, for a total of 30,000 parallel segments. The model was fine-tuned for three epochs using Low-Rank Adaptation \citep{hu_lora_2021}, a fine-tuning technique which injects small trainable matrices in the model's weights. The training corpus encompassed contemporary novels, short stories, and excerpts from science fiction and fantasy genres. The corpus was selected for its stylistic resemblance to the target text.

After fine-tuning, translation quality metrics and human inspection confirmed that Mistral-7B displayed improved handling of figurative language, idiomatic expressions, and higher accuracy. In terms of quality metrics, it achieved  +4 points of corpus-level BLEU and +7 points of COMET as compared to its off-the-shelf counterpart.

\subsection{Tools and Workflow}

To facilitate the translation and post-editing process and collect meaningful data, we used two tools: our custom-built UniOr-PET and the established PET tool~\citep{aziz_pet_2012}. 


\begin{figure}[t]
    \centering
    \includegraphics[width=\columnwidth]{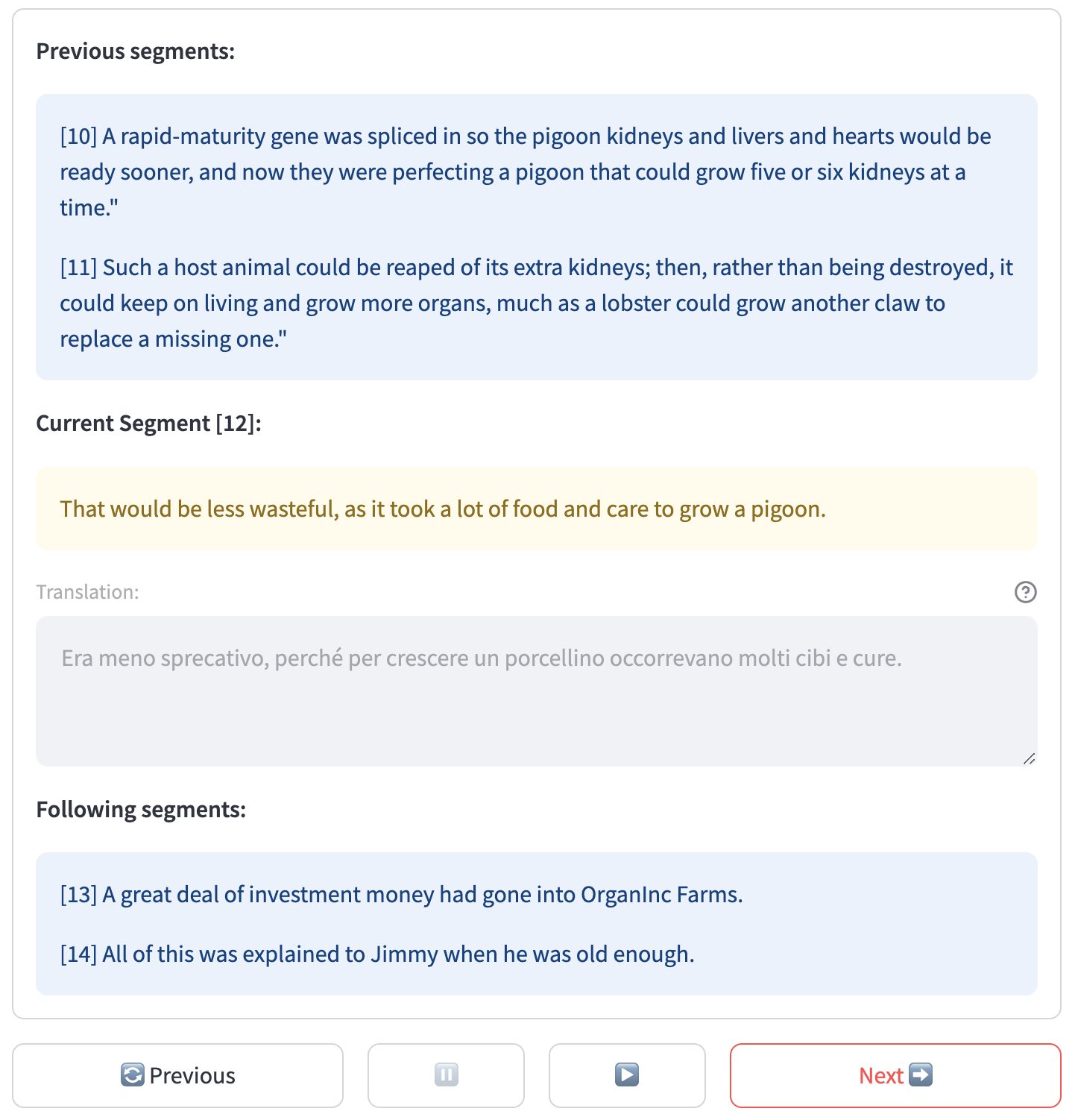}
    \caption{UniOr PET user interface}
    \label{fig:double-column}
\end{figure}

UniOr-PET was designed specifically for this study, offering a browser-based platform that eliminates the need for software installation (see Figure \ref{fig:double-column}). This feature addresses concerns often raised by translators regarding the inconvenience of downloading external applications, as is the case with the PET tool. The tool records key metrics such as editing time, the number and types of edits, keeping track of insertions, and deletions. Similarly to the PET tool, UniOr PET gives the ability to read the texts, before recording editing time, making the results from both tools equally comparable. Translators could also save their work and revisit previously edited segments. The interface was configured to present the ST, LLM output, and an editable field, with a horizontal or vertical layout.

Recognizing the importance of context in literary translation \citep{nelson_jr_literary_1989,house_text_2006}, UniOr-PET also allowed translators to view a configurable number of preceding and following segments alongside the current one. This feature ensured that they could maintain consistency in tone, style, and narrative flow, an essential consideration when translating richly detailed texts, such as literature.

In addition to UniOr-PET, translators could opt to use the PET tool, which remains a popular choice for post-editing research due to its robust functionality and familiarity among professional translators, and researchers alike. Like its browser-based relative, PET captures data such as editing times and the types of edits made, providing a rich dataset for analysis. These tools provided translators with the flexibility to choose the interface that best suited their workflow preferences while allowing us to capture detailed post-editing data.

\section{Results and Analysis}

Thanks to the use of UniOr-PET and PET, we were able to collect significant data on each translation version providing foundation for a comparative analysis of the different models. More specifically, we have calculated quality metrics with BLEU \citep{papineni_bleu_2002}, ChrF \citep{popovic_chrf_2015} and COMET \citep{rei_comet_2020}, which we average and normalize by time, as well as aggregated editing times. Finally, we compute Human-targeted Translation Edit Rate \citep{snover_study_2006}.

\subsection{Editing Times}

\begin{table}[!htbp]
\centering
\begin{tabular}{l c}
\toprule
\textbf{Source} & \textbf{Total} \\
\midrule
GPT-4 & 64.33 \\ 
Mistral-60k & 87.12 \\ 
HT & 115.68 \\ 
GPT-3.5 & 119.74 \\
\bottomrule
\end{tabular}
\caption{Editing Times (in Minutes)}
\label{tab:editing-times}
\end{table}

Table \ref{tab:editing-times} presents the aggregated total editing times (in minutes) for all translators and each part of the dataset.
We find that editing time is shorter when post-editing outputs of the larger and best performing model used in our experiment, GPT-4. Interestingly, the literary-adapted Mistral model, despite its smaller size, demonstrated editing times significantly shorter than those for GPT-3.5. This suggests that domain adaptation, even in smaller models, can have a measurable impact on post-editing efficiency. These findings align with previous research indicating that better translation quality leads to reduced post-editing effort \citep{sanchez-torron_machine_2016,zouhar_neural_2021}.

The longest editing times were recorded when translating from scratch, which is expected since it requires significantly more technical (typing) effort than post-editing pre-generated MT outputs.

\subsection{Human Translation Edit Rate (HTER)}

Table \ref{tab:hter-metrics} presents the HTER scores for the post-editing outputs from different MT systems. HTER is a widely used metric that quantifies the minimum number of edits required to improve an MT output when post-editing, where lower values indicate fewer required minimum edits. Therefore, HTER does not necessarily correspond to the actual number of edits, but rather represents an estimate of post-editing effort.

\begin{table}[!htbp]
\centering
\resizebox{\columnwidth}{!}{%
\begin{tabular}{lccccc}
\toprule
\textbf{Source} & \textbf{T1} & \textbf{T2} & \textbf{T3} & \textbf{T4} & \textbf{Doc} \\ 
\midrule
GPT-3.5 & 44.4 & 41.9 & 62.2 & 31.8 & \textbf{52} \\ 
GPT-4 & 50.4 & 66.5 & 52.2 & 29.9 & 54 \\ 
Mistral-60k & 66.1 & 66.0 & 71.5 & 54.5 & \textbf{71} \\ 
HT & 81.5 & 71.2 & 61.0 & 56.2 & 66 \\ 
\midrule
\textbf{Total} & 242.4 & 245.6 & 247.0 & 172.4 & 226.85 \\ 
\bottomrule
\end{tabular}%
}
\caption{Human Translation Edit Rate. Lowest and highest HTER values are displayed in \textbf{bold}.}
\label{tab:hter-metrics}
\end{table}

The results indicate varying levels of post-editing effort across the systems and across the four translators, with Translator 4 (T4) standing as an outlier when working with GPT models. This may be due to the adoption of a lighter form of post-editing, or an inclination to accept MT outputs considered sufficiently fluent and accurate.

We find that outputs from GPT-3.5 generally required the fewest edits, as reflected in the lowest HTER values among the systems. However, despite requiring fewer edits, post-editing outputs from GPT-3.5 took more time compared to the other models, as shown in Table \ref{tab:editing-times}. As both tools offer the possibility to read the texts, before performing translation, the results suggest that while the initial quality of GPT-3.5 translations was relatively higher, the type of edits required may have been more complex or time-consuming. 

Interestingly, GPT-4 translations required more edits than GPT-3.5 but less overall editing time, indicating that its errors were likely easier to correct. Mistral-60k, while requiring more edits than GPT-3.5 and GPT-4, had comparable or shorter editing times, possibly due to simpler or more predictable error patterns. Translations from the ST show a significant difference from the reference translation, consistent with the lack of post-editing constraints.

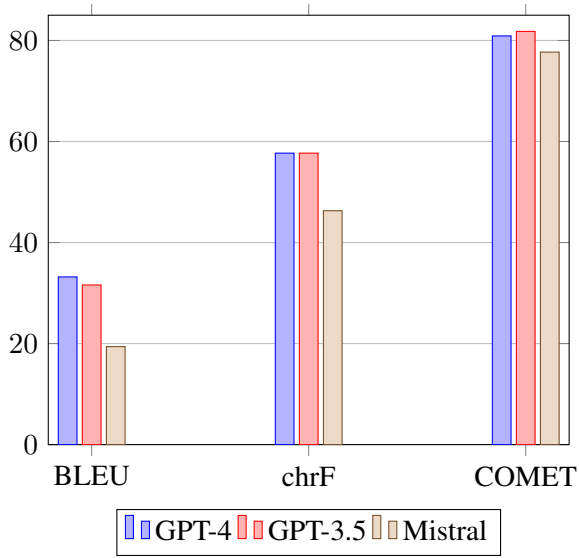
\begin{figure}[ht]
    \centering
    \begin{tikzpicture}
    \begin{axis}[
        ybar,
        bar width=7pt,
        ymin=0, ymax=85,
        symbolic x coords={BLEU, chrF, COMET},
        xtick=data,
        ymajorgrids=true,
        legend style={
            at={(0.5,-0.15)},
            anchor=north,
            legend columns=-1,
        },
    ]
    \addplot
        coordinates {(BLEU,33.2) (chrF,57.7) (COMET,80.9)};
    \addplot
        coordinates {(BLEU,31.6) (chrF,57.7) (COMET,81.8)};
    \addplot
        coordinates {(BLEU,19.4) (chrF,46.3) (COMET,77.7)};
    
    \legend{GPT-4, GPT-3.5, Mistral}
    \end{axis}
    \end{tikzpicture}
    \caption{Quality metrics scores (BLEU, chrF, COMET) for different MT systems.}
    \label{fig:quality-scores}
\end{figure}

As expected, we confirm a strong inverse correlation between HTER and quality metrics of the original MT outputs, displayed in Figure~\ref{fig:quality-scores}, indicating that lower quality MT outputs require more post-editing efforts. 

\subsection{Quality-to-Time Ratio}

Table \ref{tab:quality-time} shows the normalized quality-to-time ratio for each MT system, calculated as the average of all quality metrics (BLEU, ChrF, and COMET) divided by the total editing time (Table \ref{tab:editing-times}). This ratio provides a measure of efficiency, combining the quality of the post-edited output with the time required to achieve it. Higher values indicate more efficient systems where higher-quality translations are achieved in less time.

\begin{table}[htbp]
\centering
\resizebox{\columnwidth}{!}{%
\begin{tabular}{lcccc}
\toprule
\textbf{Source} & \textbf{Ratio} & \textbf{BLEU} & \textbf{chrF} & \textbf{COMET} \\ 
\midrule
GPT-4 & \textbf{0.38} & 31.8 & 58.2 & 83.1 \\ 
Mistral-60k & 0.29 & 27.6 & 55.0 & 83.6 \\ 
GPT-3.5 & 0.28 & 30.8 & 58.7 & 84.0 \\ 
HT & 0.23 & 27.1 & 54.4 & 80.5 \\ 
\bottomrule
\end{tabular}%
}
\caption{Quality-to-Time Ratio, calculated as the average of all quality metrics divided by the total editing time, along with BLEU, chrF, and COMET scores.}
\label{tab:quality-time}
\end{table}

The results reveal that GPT-4 achieves the highest quality-to-time ratio (0.86), demonstrating the initial quality of the translation and the reduced post-editing effort, leading to good-quality post-edited translations in the shortest time.

Interestingly, Mistral-60k achieves the lowest ratio across the three models, despite requiring less editing time compared to GPT-3.5. This suggests that while Mistral translations may be quicker to edit, their initial quality presents challenges that limit their effectiveness in producing high-quality outputs efficiently, possibly resulting in the translator's decision to perform a lighter form of post-editing \citep{nitzke_short_2021}.

\section{Creativity Annotation}

To evaluate creativity in the post-edited translations and conduct a model-wise comparison, we annotated units of creative potential in the ST and creative shifts in the target texts (TT), that were originally generated by the three LLMs, and then post-edited by four translators. 

\begin{figure*}[!ht]
\[
\text{Creativity Score}
= \Bigl(
   \frac{\#\text{CSs}}{\#\text{UCPs}}
   \;-\;
   \frac{\#\text{error points} - \#\text{Kudos}}{\#\text{words in ST}}
\Bigr)
\times 100.
\]
\caption{The original creativity score formula, that we started from to create our score.}
\label{fig:creativity-score}
\end{figure*}

\paragraph{Annotation Process.}
Our annotation framework follows the methodology proposed by \citet{guerberof-arenas_creativity_2022}, where units of creative potential (UCPs) are defined as units that could invite creative deviations during post-editing, aimed at preserving or enhancing the creativity found in the ST, and creative shifts reflect the actual creative units introduced by translators during post-editing. Annotations were performed by two linguists with expertise in translation studies, who are native speakers of the target language and proficient in English. After annotating 10\% of the dataset, inter-annotator agreement (IAA) was calculated to ensure the reliability of the annotations. The initial agreement, measured with Cohen's Kappa, was equal to $K=0.35$ for Type Agreement and $K=0.85$ for Span Agreement, due to disagreements primarily on the type of creative shift to assign, rather than the identification of the creative shifts themselves. Following a collaborative resolution process, we refined the annotation guidelines and calculated agreement on the final annotations, reaching a Type Agreement equal to $K=0.57$ and a similarly high Span Agreement, equal to $K=0.86$.

\paragraph{Creativity Score.}
A creative work must be both novel and acceptable, thereby achieving a balance between creativity and quality. In order to account for both novelty, as indicated by the number of creative shifts, and acceptability, as reflected by translation quality, we used \texttt{WMT22-COMET-DA} \citep{rei_comet-22_2022} for an automatic reference-based quality evaluation, and calculated the creativity score across the four translations.

In this study, we employ COMET as our primary metric for assessing translation quality, recognizing that MQM would provide a more fine-grained evaluation of translation errors. Our decision to use COMET is motivated by its strong correlation with human judgments, as demonstrated in previous research \citep{rei_comet_2020,kocmi_navigating_2024}, and by its practical advantage in automatic evaluation, in light of constraints related to time and resources. Having been trained on MQM-annotated datasets, COMET should effectively reflect the types of errors found in the outputs. Therefore, we integrate COMET in our creativity evaluation formula, as a proxy for translation acceptability.

Compared to the formula used in the original study, presented in Figure~\ref{fig:creativity-score}, we adapt the acceptability equation to accommodate the use of a quality metric, where higher means better, in place of the original error metric. Therefore, we multiply the creative shifts ratio by COMET scores, and then multiply by 100 to express it as a percentage. This allows us to reward creativity in proportion to quality, similarly to the original study. We present the new creativity score formula below.

\begin{equation}
\label{eq:creativity-new-score}
\text{CS} 
= \left( \frac{\textit{Creative Shifts}}{\textit{UCPs}} \times \text{COMET} \right) \times 100
\end{equation}

\subsection {Annotation Results}
Table~\ref{tab:cs-annotation} summarizes the annotation results for each translation variant. For each system, we present the number of the creative shifts introduced by the translators, the COMET score, and the resulting creativity score, calculated with our new formula. A higher creativity score suggests a better balance between the introduced creative elements and the final translation quality. 


\begin{table}[h]
    \centering
    \resizebox{\columnwidth}{!}{%
    \begin{tabular}{lccc}
        \toprule
        \textbf{System} & \textbf{CS Ratio} & \textbf{COMET} & \textbf{Creativity} \\
        \midrule
        HT & 0.30 & \textbf{0.85} & 25.5\% \\
        GPT-3.5 & 0.24 & 0.84 & 20.1\% \\
        Mistral & 0.30 & 0.83 & 24.9\% \\
        GPT-4 & \textbf{0.32} & 0.83 & \textbf{26.5\%} \\
        \bottomrule
    \end{tabular}%
}
    \caption{Creativity annotation results, where we display Creative Shifts ratio, COMET Score, and Creativity Score for each system.}
    \label{tab:cs-annotation}
\end{table}

\section{Discussion}


Taken together, our results show that a larger and more advanced model (GPT-4) generated translations that required fewer edits and resulted in a higher-quality post-edited translation, as resulted from the lower editing time and the higher quality-to-time ratio. The creativity score is also the highest, suggesting an interesting correlation between original MT quality and creativity in post-editing.

The domain-adapted Mistral-7B model also displayed promising performance, obtaining a quality-to-time ratio higher than the one obtained by the larger GPT-3.5, requiring more edits but a significantly lower editing time, while obtaining a similar creativity score. In this case, we find that Mistral's creativity comes at the cost of increased post-editing effort. HT, despite requiring a significantly higher editing time, is the most accurate translation variant according to COMET scores and it presents a high creativity score that is very similar to the post-edited texts. 

In Table~\ref{tab:hter-examples} we present two segments for each translation version with the highest and lowest post-editing effort, as measured by HTER. In displaying the segments, we ignore cases where the HTER is equal to zero due to translators not making any changes to the MT output. The examples reveal several interesting patterns. In some cases, the translators decided to merge or split certain sentences. Extensive edits were made in segments containing UCPs, as in the second example for GPT-3.5. Similarly, we find several edits where the original MT quality was particularly low, as seen in the second segment from the Mistral model. Interestingly, we find that where the MT systems failed to render neologisms effectively, translators were forced to produce a creative alternative, effectively improving the creativity of the translation.

\begin{table*}[ht]
\centering
\small
\renewcommand{\arraystretch}{1.2}
\begin{tabular}{p{2.5cm}|p{0.8cm}|p{10.5cm}}
\toprule
\textbf{Model} & \textbf{Type} & \textbf{Text} \\
\midrule

\multirow{4}{*}{
  \begin{tabular}[c]{@{}l@{}}
    \textbf{GPT-3.5} \\
    \textit{(Lowest HTER)}
  \end{tabular}
}
  & ST & \textit{But he hadn't wet his bed for a long time, or he didn't think he had.} \\
  & HT & \textit{Eppure era un pezzo che non bagnava il letto, o almeno così credeva.} \\
  & MT & \textit{Ma non aveva bagnato il letto da molto tempo, o almeno non pensava di averlo fatto.} \\
  & \textbf{PE} & \textit{Eppure era da un pezzo che non bagnava il letto, o almeno così credeva.} \\

\midrule

\multirow{4}{*}{
  \begin{tabular}[c]{@{}l@{}}
    \textbf{GPT-3.5} \\
    \textit{(Highest HTER)}
  \end{tabular}
}
  & ST & \textit{Some cheap do-it-yourself enlightenment handbook, Nirvana for halfwits.} \\
  & HT & \textit{Uno scadente manuale di auto rivelazione per gonzi.} \\
  & MT & \textit{Una specie di manuale economico per l'illuminazione fai-da-te (...)} \\
  & \textbf{PE} & \textit{Una specie di manuale a poco prezzo per raggiungere l'illuminazione (...)} \\

\midrule
\midrule

\multirow{4}{*}{
  \begin{tabular}[c]{@{}l@{}}
    \textbf{GPT-4o} \\
    \textit{(Lowest HTER)}
  \end{tabular}
}
  & ST & \textit{All of this was explained to Jimmy when he was old enough.
} \\
  & HT & \textit{Tutto questo fu spiegato a Jimmy quando fu abbastanza grande.} \\
  & MT & \textit{Tutto questo fu spiegato a Jimmy quando era abbastanza grande.} \\
  & \textbf{PE} & \textit{Tutto questo venne spiegato a Jimmy quando fu abbastanza grande.} \\

\midrule

\multirow{4}{*}{
  \begin{tabular}[c]{@{}l@{}}
    \textbf{GPT-4o} \\
    \textit{(Highest HTER)}
  \end{tabular}
}
  & ST & \textit{She's got her own ideas.} \\
  & HT & \textit{Ha le sue idee.} \\
  & MT & \textit{He le sue proprie idee.} \\
  & \textbf{PE} & \textit{Abbiamo opinioni diverse sulla cosa.} \\

\midrule
\midrule

\multirow{4}{*}{
  \begin{tabular}[c]{@{}l@{}}
    \textbf{Mistral} \\
    \textit{(Lowest HTER)}
  \end{tabular}
}
  & ST & \textit{Ramona was one of his dad's lab technicians.} \\
  & HT & \textit{Ramona era uno dei tecnici di laboratorio di suo padre.} \\
  & MT & \textit{Ramona era una delle tecniche del laboratorio del padre.} \\
  & \textbf{PE} & \textit{Ramona era una dei tecnici del laboratorio di suo padre.} \\

\midrule

\multirow{4}{*}{
  \begin{tabular}[c]{@{}l@{}}
    \textbf{Mistral} \\
    \textit{(Highest HTER)}
  \end{tabular}
}
  & ST & \textit{They called the cities the pleeblands.} \\
  & HT & \textit{Chiamavano le città plebopoli.} \\
  & MT & \textit{Chiamavano le città le plebe.} \\
  & \textbf{PE} & \textit{Si riferivano alle città chiamandole terre di plebelandia.} \\

\midrule
\bottomrule
\end{tabular}
\caption{Examples of source text (ST), human translation (HT), machine translation (MT), and post-edited output (PE) for GPT-4o, GPT-3.5, and Mistral, showing segments with the lowest and highest post-editing effort as measured by HTER.}
\label{tab:hter-examples}
\end{table*}

Overall, we find that the creativity score does not differ significantly between the four models, as both the number of identified creative shifts and the quality metrics are similar across all translation variants. These findings are in contrast with what was found in the original study, where the difference between the two modalities (HT and PE) was substantial and HT was found to be notably more creative than their post-edited variant. We speculate that the higher and more fluent MT quality given by LLMs may be of less constraint to the translator in the post-editing process, leading to equally creative translations.

\section{Conclusion}

In this study, we investigated the potential of LLM-based post-editing in the literary domain, comparing a literary-adapted Mistral model with GPT-4 and GPT-3.5. By collaborating with four professional literary translators, we collected detailed data on editing times, error rates, and post-editing efficiency, using our custom-built tool UniOr-PET. We demonstrate the contributions that LLMs can make in literary post-editing workflows, bridging the gap between productivity and creativity. 

Our findings highlight two important benefits granted by the adoption of LLMs. First, we demonstrate that, in the context of our study, creativity does not present a significant difference between human translation and post-edited LLM translations. The marginal difference in creativity between the four translation variants suggests that the post-edited outputs may preserve creativity effectively. This may be due to the more fluent and higher-quality outputs given by the original MT versions, that represent less of a constraint to the translators, compared to NMT outputs.

Second, we observe a clear productivity gain in post-editing compared to human translation, even when post-editing translations generated by a smaller model. Given that the creativity gap is relatively small across translation variants, the productivity gains may offset the minor differences in creativity, achieving similarly creative translations with significantly less effort and time.

Finally, we reinforce the potential of fine-tuning techniques for literary MT workflows, demonstrating that even by adopting a small literary-adapted model, it is possible to achieve a good balance between translation quality and efficiency.  

\section{Limitations}

One of the main limitations of this study is that our data collection process involved only four translators working in a single relatively well-resourced language pair and a relatively short literary excerpt. Further studies, on a larger scale, are required to investigate the possible correlations between creativity and other metrics. It is also worth mentioning that although our study follows established proxies for measuring creativity, these should be verified with a reception study, as suggested by \citet{guerberof-arenas_impact_2020}.

For the acceptability score, meant to balance creativity by translation quality in the post-edited texts, we used COMET scores in place of human evaluation. While COMET has shown strong correlations with human judgment, it remains an automated metric and may not fully capture the extent of literary translation quality.

Finally, while our literary-adapted Mistral model showed promising performance, its fine-tuning was performed using a modest-sized corpus, leaving open the way for further experimentation.

\subsection{CO2 Emission Related to Experiments}

Experiments were conducted using Amazon Web Services in region eu-west-1, which has a carbon efficiency of 0.62 kgCO$_2$eq/kWh. A cumulative of 3 hours of computation was performed on hardware of type RTX A6000 (TDP of 300W).

Total emissions are estimated to be 0.56 kgCO$_2$eq of which 100 percents were directly offset by the cloud provider.
    
Estimations were conducted using the \href{https://mlco2.github.io/impact#compute}{Machine Learning Impact calculator} presented in \citet{lacoste2019quantifying}.



\bibliography{references}

\begin{thebibliography}{36}
\providecommand{\natexlab}[1]{#1}

\bibitem[{Atwood(2004)}]{atwood_oryx_2004}
Margaret Atwood. 2004.
\newblock \emph{Oryx and {Crake}}.
\newblock Number v.1 in The {MaddAddam} {Trilogy} {Ser}. Knopf Doubleday
  Publishing Group, New York.

\bibitem[{Aziz et~al.(2012)Aziz, Castilho, and Specia}]{aziz_pet_2012}
Wilker Aziz, Sheila Castilho, and Lucia Specia. 2012.
\newblock \href {https://aclanthology.org/L12-1587/} {{PET}: a {Tool} for
  {Post}-editing and {Assessing} {Machine} {Translation}}.
\newblock In \emph{Proceedings of the {Eighth} {International} {Conference} on
  {Language} {Resources} and {Evaluation} ({LREC}`12)}, pages 3982--3987,
  Istanbul, Turkey. European Language Resources Association (ELRA).

\bibitem[{Brown et~al.(2020)Brown, Mann, Ryder, Subbiah, Kaplan, Dhariwal,
  Neelakantan, Shyam, Sastry, Askell, Agarwal, Herbert-Voss, Krueger, Henighan,
  Child, Ramesh, Ziegler, Wu, Winter, Hesse, Chen, Sigler, Litwin, Gray, Chess,
  Clark, Berner, McCandlish, Radford, Sutskever, and
  Amodei}]{brown_language_2020}
Tom~B. Brown, Benjamin Mann, Nick Ryder, Melanie Subbiah, Jared Kaplan,
  Prafulla Dhariwal, Arvind Neelakantan, Pranav Shyam, Girish Sastry, Amanda
  Askell, Sandhini Agarwal, Ariel Herbert-Voss, Gretchen Krueger, Tom Henighan,
  Rewon Child, Aditya Ramesh, Daniel~M. Ziegler, Jeffrey Wu, Clemens Winter,
  Christopher Hesse, Mark Chen, Eric Sigler, Mateusz Litwin, Scott Gray,
  Benjamin Chess, Jack Clark, Christopher Berner, Sam McCandlish, Alec Radford,
  Ilya Sutskever, and Dario Amodei. 2020.
\newblock Language models are few-shot learners.
\newblock In \emph{Proceedings of the 34th {International} {Conference} on
  {Neural} {Information} {Processing} {Systems}}, {NIPS} '20, pages 1877--1901,
  Red Hook, NY, USA. Curran Associates Inc.

\bibitem[{Castaldo et~al.(2025)Castaldo, Castilho, Moorkens, and
  Monti}]{castaldo_2025_UniOrPET}
Antonio Castaldo, Sheila Castilho, Joss Moorkens, and Johanna Monti. 2025.
\newblock \href {https://mtsummit2025.unige.ch/} {Unior {PET}: {An} {Online}
  {Platform} for {Translation} {Post-Editing}}.
\newblock In \emph{20th Machine Translation Summit: {Products and Projects
  track}}, Geneva, Switzerland. European Association for Machine Translation.

\bibitem[{Castaldo and Monti(2024)}]{castaldo_prompting_2024}
Antonio Castaldo and Johanna Monti. 2024.
\newblock \href {https://unora.unior.it/handle/11574/231020} {Prompting {Large}
  {Language} {Models} for {Idiomatic} {Translation}}.
\newblock In \emph{Proceedings of the {First} {Workshop} on {Creative}-text
  {Translation} and {Technology}}, pages 37--44, Sheffield, UK.
\newblock Accepted: 2024-06-19T21:00:05Z.

\bibitem[{Castilho et~al.(2023)Castilho, Mallon, Meister, and
  Yue}]{castilho_online_2023}
Sheila Castilho, Clodagh~Quinn Mallon, Rahel Meister, and Shengya Yue. 2023.
\newblock \href {https://aclanthology.org/2023.eamt-1.39} {Do online {Machine}
  {Translation} {Systems} {Care} for {Context}? {What} {About} a {GPT}
  {Model}?}
\newblock In \emph{Proceedings of the 24th {Annual} {Conference} of the
  {European} {Association} for {Machine} {Translation}}, pages 393--417,
  Tampere, Finland. European Association for Machine Translation.

\bibitem[{Castilho and Resende(2022)}]{castilho_post-editese_2022}
Sheila Castilho and Natália Resende. 2022.
\newblock \href {https://doi.org/10.3390/info13020066} {Post-{Editese} in
  {Literary} {Translations}}.
\newblock \emph{Information}, 13(2):66.
\newblock Number: 2 Publisher: Multidisciplinary Digital Publishing Institute.

\bibitem[{Corpas~Pastor and
  Noriega-Santiáñez(2024)}]{corpas_pastor_human_2024}
Gloria Corpas~Pastor and Laura Noriega-Santiáñez. 2024.
\newblock \href {https://doi.org/10.3390/info15090530} {Human versus {Neural}
  {Machine} {Translation} {Creativity}: {A} {Study} on {Manipulated} {MWEs} in
  {Literature}}.
\newblock \emph{Information}, 15(9):530.

\bibitem[{Guerberof~Arenas(2014)}]{guerberof_arenas_correlations_2014}
Ana Guerberof~Arenas. 2014.
\newblock \href {https://doi.org/10.1007/s10590-014-9155-y} {Correlations
  between productivity and quality when post-editing in a professional
  context}.
\newblock \emph{Machine Translation}, 28(3):165--186.

\bibitem[{Guerberof-Arenas and Toral(2020)}]{guerberof-arenas_impact_2020}
Ana Guerberof-Arenas and Antonio Toral. 2020.
\newblock \href {https://doi.org/10.1075/ts.20035.gue} {The impact of
  post-editing and machine translation on creativity and reading experience}.
\newblock \emph{Translation Spaces}, 9(2):255--282.

\bibitem[{Guerberof-Arenas and Toral(2022)}]{guerberof-arenas_creativity_2022}
Ana Guerberof-Arenas and Antonio Toral. 2022.
\newblock \href {https://doi.org/10.1075/ts.21025.gue} {Creativity in
  translation: {Machine} translation as a constraint for literary texts}.
\newblock \emph{Translation Spaces}, 11(2):184--212.

\bibitem[{Gurov(2022)}]{gurov_literary_2022}
Andrey Gurov. 2022.
\newblock \href {https://doi.org/10.2139/ssrn.4299110} {Literary {Translation}
  as {An} {Insurmountable} {Obstacle} for {Neural} {Networks}}.
\newblock \emph{SSRN Electronic Journal}.

\bibitem[{House(2006)}]{house_text_2006}
Juliane House. 2006.
\newblock \href {https://doi.org/10.1016/j.pragma.2005.06.021} {Text and
  context in translation}.
\newblock \emph{Journal of Pragmatics}, 38(3):338--358.

\bibitem[{Hu et~al.(2021)Hu, Shen, Wallis, Allen-Zhu, Li, Wang, Wang, and
  Chen}]{hu_lora_2021}
Edward~J. Hu, Yelong Shen, Phillip Wallis, Zeyuan Allen-Zhu, Yuanzhi Li, Shean
  Wang, Lu~Wang, and Weizhu Chen. 2021.
\newblock \href {https://doi.org/10.48550/arXiv.2106.09685} {{LoRA}:
  {Low}-{Rank} {Adaptation} of {Large} {Language} {Models}}.
\newblock \emph{arXiv preprint}.
\newblock ArXiv:2106.09685 [cs].

\bibitem[{Jiang et~al.(2023)Jiang, Sablayrolles, Mensch, Bamford, Chaplot,
  Casas, Bressand, Lengyel, Lample, Saulnier, Lavaud, Lachaux, Stock, Scao,
  Lavril, Wang, Lacroix, and Sayed}]{jiang_mistral_2023}
Albert~Q. Jiang, Alexandre Sablayrolles, Arthur Mensch, Chris Bamford,
  Devendra~Singh Chaplot, Diego de~las Casas, Florian Bressand, Gianna Lengyel,
  Guillaume Lample, Lucile Saulnier, Lélio~Renard Lavaud, Marie-Anne Lachaux,
  Pierre Stock, Teven~Le Scao, Thibaut Lavril, Thomas Wang, Timothée Lacroix,
  and William~El Sayed. 2023.
\newblock \href {https://doi.org/10.48550/arXiv.2310.06825} {Mistral {7B}}.
\newblock \emph{arXiv preprint}.
\newblock Issue: arXiv:2310.06825 arXiv:2310.06825 [cs].

\bibitem[{Karpinska and Iyyer(2023)}]{karpinska_large_2023}
Marzena Karpinska and Mohit Iyyer. 2023.
\newblock \href {https://doi.org/10.48550/arXiv.2304.03245} {Large language
  models effectively leverage document-level context for literary translation,
  but critical errors persist}.
\newblock Issue: arXiv:2304.03245 arXiv:2304.03245 [cs].

\bibitem[{Kocmi et~al.(2024)Kocmi, Zouhar, Federmann, and
  Post}]{kocmi_navigating_2024}
Tom Kocmi, Vilém Zouhar, Christian Federmann, and Matt Post. 2024.
\newblock \href {https://doi.org/10.48550/arXiv.2401.06760} {Navigating the
  {Metrics} {Maze}: {Reconciling} {Score} {Magnitudes} and {Accuracies}}.
\newblock Issue: arXiv:2401.06760 arXiv:2401.06760 [cs].

\bibitem[{Lacoste et~al.(2019)Lacoste, Luccioni, Schmidt, and
  Dandres}]{lacoste2019quantifying}
Alexandre Lacoste, Alexandra Luccioni, Victor Schmidt, and Thomas Dandres.
  2019.
\newblock Quantifying the carbon emissions of machine learning.
\newblock \emph{arXiv preprint arXiv:1910.09700}.

\bibitem[{Miller(2019)}]{miller_punsters_2019}
Tristan Miller. 2019.
\newblock \href {https://doi.org/10.26615/issn.2683-0078.2019_007} {The
  {Punster}`s {Amanuensis}: {The} {Proper} {Place} of {Humans} and {Machines}
  in the {Translation} of {Wordplay}}.
\newblock In \emph{Proceedings of the {Human}-{Informed} {Translation} and
  {Interpreting} {Technology} {Workshop} ({HiT}-{IT} 2019)}, pages 57--65,
  Varna, Bulgaria. Incoma Ltd., Shoumen, Bulgaria.

\bibitem[{Moorkens et~al.(2018)Moorkens, Toral, Castilho, and
  Way}]{moorkens_translators_2018}
Joss Moorkens, Antonio Toral, Sheila Castilho, and Andy Way. 2018.
\newblock \href {https://doi.org/10.1075/ts.18014.moo} {Translators’
  perceptions of literary post-editing using statistical and neural machine
  translation}.
\newblock \emph{Translation Spaces}, 7(2):240--262.
\newblock Publisher: John Benjamins Publishing Company.

\bibitem[{Moslem et~al.(2023)Moslem, Haque, Kelleher, and
  Way}]{moslem_adaptive_2023}
Yasmin Moslem, Rejwanul Haque, John~D. Kelleher, and Andy Way. 2023.
\newblock \href {https://aclanthology.org/2023.eamt-1.22} {Adaptive {Machine}
  {Translation} with {Large} {Language} {Models}}.
\newblock In \emph{Proceedings of the 24th {Annual} {Conference} of the
  {European} {Association} for {Machine} {Translation}}, pages 227--237,
  Tampere, Finland. European Association for Machine Translation.

\bibitem[{Nelson~Jr.(1989)}]{nelson_jr_literary_1989}
Lowry Nelson~Jr. 1989.
\newblock \href {https://doi.org/10.1080/07374836.1989.10523445} {Literary
  {Translation}}.
\newblock \emph{Translation Review}, 29(1):17--30.
\newblock Publisher: Routledge.

\bibitem[{Nitzke and Hansen-Schirra(2021)}]{nitzke_short_2021}
Jean Nitzke and Silvia Hansen-Schirra. 2021.
\newblock \href {https://library.oapen.org/handle/20.500.12657/52585} {\emph{A
  short guide to post-editing ({Volume} 16)}}.
\newblock Language Science Press.

\bibitem[{Noriega-Santiáñez and
  Corpas~Pastor(2023)}]{noriega-santianez_machine_2023}
Laura Noriega-Santiáñez and Gloria Corpas~Pastor. 2023.
\newblock \href {https://doi.org/10.5565/rev/tradumatica.338} {Machine vs
  {Human} {Translation} of {Formal} {Neologisms} in {Literature}: {Exploring}
  {E}-tools and {Creativity} in {Students}}.
\newblock \emph{Tradumàtica tecnologies de la traducció}, (21):233--264.

\bibitem[{OpenAI et~al.(2024)OpenAI, Achiam, Adler, Agarwal, Ahmad, Akkaya,
  Aleman, Almeida, Altenschmidt, Altman, Belgum, Bello, Berdine,
  Bernadett-Shapiro, Berner, Bogdonoff, Boiko, Boyd, Brakman, and
  Brockman}]{openai_gpt-4_2024}
OpenAI, Josh Achiam, Steven Adler, Sandhini Agarwal, Lama Ahmad, Ilge Akkaya,
  Florencia~Leoni Aleman, Diogo Almeida, Janko Altenschmidt, Sam Altman, Jeff
  Belgum, Irwan Bello, Jake Berdine, Gabriel Bernadett-Shapiro, Christopher
  Berner, Lenny Bogdonoff, Oleg Boiko, Madelaine Boyd, Anna-Luisa Brakman, and
  Greg Brockman. 2024.
\newblock \href {https://doi.org/10.48550/arXiv.2303.08774} {{GPT}-4
  {Technical} {Report}}.
\newblock \emph{arXiv preprint}.
\newblock ArXiv:2303.08774 [cs].

\bibitem[{Papineni et~al.(2002)Papineni, Roukos, Ward, and
  Zhu}]{papineni_bleu_2002}
Kishore Papineni, Salim Roukos, Todd Ward, and Wei-Jing Zhu. 2002.
\newblock \href {https://doi.org/10.3115/1073083.1073135} {Bleu: a {Method} for
  {Automatic} {Evaluation} of {Machine} {Translation}}.
\newblock In \emph{Proceedings of the 40th {Annual} {Meeting} of the
  {Association} for {Computational} {Linguistics}}, pages 311--318,
  Philadelphia, Pennsylvania, USA. Association for Computational Linguistics.

\bibitem[{Popović(2015)}]{popovic_chrf_2015}
Maja Popović. 2015.
\newblock \href {https://doi.org/10.18653/v1/W15-3049} {{chrF}: character
  n-gram {F}-score for automatic {MT} evaluation}.
\newblock In \emph{Proceedings of the {Tenth} {Workshop} on {Statistical}
  {Machine} {Translation}}, pages 392--395, Lisbon, Portugal. Association for
  Computational Linguistics.

\bibitem[{Raunak et~al.(2023)Raunak, Menezes, Post, and
  Awadalla}]{raunak_gpts_2023}
Vikas Raunak, Arul Menezes, Matt Post, and Hany~Hassan Awadalla. 2023.
\newblock \href {http://arxiv.org/abs/2305.16806} {Do {GPTs} {Produce} {Less}
  {Literal} {Translations}?}
\newblock ArXiv: 2305.16806.

\bibitem[{Rei et~al.(2022)Rei, C.~de Souza, Alves, Zerva, Farinha, Glushkova,
  Lavie, Coheur, and Martins}]{rei_comet-22_2022}
Ricardo Rei, José~G. C.~de Souza, Duarte Alves, Chrysoula Zerva, Ana~C
  Farinha, Taisiya Glushkova, Alon Lavie, Luisa Coheur, and André F.~T.
  Martins. 2022.
\newblock \href {https://aclanthology.org/2022.wmt-1.52} {{COMET}-22:
  {Unbabel}-{IST} 2022 {Submission} for the {Metrics} {Shared} {Task}}.
\newblock In \emph{Proceedings of the {Seventh} {Conference} on {Machine}
  {Translation} ({WMT})}, pages 578--585, Abu Dhabi, United Arab Emirates
  (Hybrid). Association for Computational Linguistics.

\bibitem[{Rei et~al.(2020)Rei, Stewart, Farinha, and Lavie}]{rei_comet_2020}
Ricardo Rei, Craig Stewart, Ana~C Farinha, and Alon Lavie. 2020.
\newblock \href {https://doi.org/10.18653/v1/2020.emnlp-main.213} {{COMET}: {A}
  {Neural} {Framework} for {MT} {Evaluation}}.
\newblock In \emph{Proceedings of the 2020 {Conference} on {Empirical}
  {Methods} in {Natural} {Language} {Processing} ({EMNLP})}, pages 2685--2702,
  Online. Association for Computational Linguistics.

\bibitem[{Runco and Jaeger(2012)}]{runco_standard_2012}
Mark~A. Runco and Garrett~J. Jaeger. 2012.
\newblock \href {https://doi.org/10.1080/10400419.2012.650092} {The {Standard}
  {Definition} of {Creativity}}.
\newblock \emph{Creativity Research Journal}, 24(1):92--96.

\bibitem[{Sanchez-Torron and Koehn(2016)}]{sanchez-torron_machine_2016}
Marina Sanchez-Torron and Philipp Koehn. 2016.
\newblock \href {https://aclanthology.org/2016.amta-researchers.2/} {Machine
  {Translation} {Quality} and {Post}-{Editor} {Productivity}}.
\newblock In \emph{Conferences of the {Association} for {Machine} {Translation}
  in the {Americas}: {MT} {Researchers}' {Track}}, pages 16--26, Austin, TX,
  USA. The Association for Machine Translation in the Americas.

\bibitem[{Snover et~al.(2006)Snover, Dorr, Schwartz, Micciulla, and
  Makhoul}]{snover_study_2006}
Matthew Snover, Bonnie Dorr, Rich Schwartz, Linnea Micciulla, and John Makhoul.
  2006.
\newblock \href {https://aclanthology.org/2006.amta-papers.25/} {A {Study} of
  {Translation} {Edit} {Rate} with {Targeted} {Human} {Annotation}}.
\newblock In \emph{Proceedings of the 7th {Conference} of the {Association} for
  {Machine} {Translation} in the {Americas}: {Technical} {Papers}}, pages
  223--231, Cambridge, Massachusetts, USA. Association for Machine Translation
  in the Americas.

\bibitem[{Terribile(2023)}]{terribile_is_2023}
Silvia Terribile. 2023.
\newblock \href {https://doi.org/10.1075/ts.22044.ter} {Is post-editing really
  faster than human translation?}
\newblock \emph{Translation Spaces}, 13(2):171--199.
\newblock Publisher: John Benjamins Publishing Company.

\bibitem[{Tiedemann and Thottingal(2020)}]{tiedemann_opus-mt_2020}
Jörg Tiedemann and Santhosh Thottingal. 2020.
\newblock \href {https://aclanthology.org/2020.eamt-1.61} {{OPUS}-{MT} –
  {Building} open translation services for the {World}}.
\newblock In \emph{Proceedings of the 22nd {Annual} {Conference} of the
  {European} {Association} for {Machine} {Translation}}, pages 479--480,
  Lisboa, Portugal. European Association for Machine Translation.

\bibitem[{Zouhar et~al.(2021)Zouhar, Popel, Bojar, and
  Tamchyna}]{zouhar_neural_2021}
Vilém Zouhar, Martin Popel, Ondřej Bojar, and Aleš Tamchyna. 2021.
\newblock \href {https://doi.org/10.18653/v1/2021.emnlp-main.801} {Neural
  {Machine} {Translation} {Quality} and {Post}-{Editing} {Performance}}.
\newblock In \emph{Proceedings of the 2021 {Conference} on {Empirical}
  {Methods} in {Natural} {Language} {Processing}}, pages 10204--10214, Online
  and Punta Cana, Dominican Republic. Association for Computational
  Linguistics.

\end{thebibliography}

\end{document}